%% file: main.tex
\documentclass[10pt,twocolumn,letterpaper]{article}

\usepackage[pagenumbers]{cvpr} % To force page numbers, e.g. for an arXiv version
\usepackage{breqn}
\input{preamble}

\definecolor{cvprblue}{rgb}{0.21,0.49,0.74}
\usepackage[pagebackref,breaklinks,colorlinks,citecolor=cvprblue]{hyperref}

\def\paperID{*****} % *** Enter the Paper ID here
\def\confName{CVPR}
\def\confYear{2024}

\title{Technical Report for CVPR 2024 5th CLVISION Challenge:\\
Multi-Level Knowledge Distillation and Dynamic Self-Supervised Learning for Continual Learning.
}

\author{
Taeheon Kim$^{1,2}$\hspace{0.5em}
San Kim$^{1,2}$\hspace{0.5em}
Minhyuk Seo$^{1,2}$\hspace{0.5em}
Dongjae Jeon$^{1,2}$\hspace{0.5em}
Wonje Jeung$^{1,2}$\vspace{0.5em}
Jonghyun Choi$^{2}$\vspace{0.5em}\\
{\hspace{0.5em}$^1$Yonsei University\hspace{0.5em}$^2$Seoul National University} \vspace{0.5em}\\
\small\texttt{\{thkim0305,nasmik419,dbd0508,dongjae0324,specific0924\}@yonsei.ac.kr} \\
\small\texttt{jonghyunchoi@snu.ac.kr}\\}

\begin{document}
\maketitle
\input{sec/0_abstract}

\input{sec/1_intro}

\input{sec/2_method}
\input{sec/3_experiment}
\input{sec/4_concolusion}
{
    \small
    \bibliographystyle{ieeenat_fullname}
    \bibliography{main}
}


\end{document}

%% file: preamble.tex
\usepackage[dvipsnames]{xcolor}

%% file: sec/0_abstract.tex
\begin{abstract}
Class-incremental with repetition (CIR), where previously trained classes repeatedly introduced in future tasks, is a more realistic scenario than the traditional class incremental setup, which assumes that each task contains unseen classes.
% The primary objective of CIR is to ensure high stability and plasticity in models, enabling them to perform effectively on learned tasks without causing catastrophic forgetting. 
% Additionally, in real-world applications of CIR, we can easily acquire abundant unlabeled data from external sources such as the Internet.
% Thus, we propose two components that utilize the unlabeled data to enhance the plasticity and the stability of the network.  
CIR assumes that we can easily access abundant unlabeled data from external sources, such as the Internet.
Therefore, we propose two components that efficiently use the unlabeled data to ensure the high stability and the plasticity of models trained in CIR setup.
First, we introduce multi-level knowledge distillation (MLKD) that distills knowledge from multiple previous models across multiple perspectives, including features and logits, so the model can maintain much various previous knowledge.
Moreover, we implement dynamic self-supervised loss (SSL) to utilize the unlabeled data that accelerates the learning of new classes, while dynamic weighting of SSL keeps the focus of training to the primary task. 
Both of our proposed components significantly improve the performance in CIR setup, achieving 2nd place in the CVPR 5th CLVISION Challenge\footnote{Code is available at \url{https://github.com/ta3h30nk1m/5th_CLVISION_Challenge_MLKD_SSL}}.
\end{abstract}

%% file: sec/1_intro.tex
\section{Introduction}
\label{sec:intro}

% Class-Incremental Learning (CIL) is a crucial paradigm in deep learning that enables models to continuously acquire and integrate new knowledge, such as recognizing new object categories, without losing previously learned information~\cite{rebuffi2017icarl}. 
Class incremental learning is a setup where the model learns new classes while preventing forgetting of the previously seen classes.
Traditionally, CIL setups assume that new classes are introduced with each new task and that a fixed number of new classes are added per task.
% However, this assumption has been critiqued as unrealistic by Hemati et al.~\cite{hemati2023class}, who argue that in practical scenarios, data may be introduced sequentially without always containing new classes.
% To address this limitation, they proposed Class-Incremental Learning with Repetition (CIR), a more flexible framework where the model encounters a mixture of old and new classes throughout the learning process.
However, in a real-world scenario, a model may encounter both new and previously seen classes. 
To bridge the gap between these scenarios and the existing CIL setup, \citet{hemati2023class} proposed Class-Incremental Learning with Repetition (CIR), a more flexible framework where the model encounters a mixture of old and new classes.
This setup aligns with concepts from other learning paradigms, such as the non-IID assumption in federated learning\cite{karimireddy2020scaffold} and the blurry task boundaries in online continual learning~\cite{koh2022online}, challenging the applicability of methods that rely on stricter assumptions.

% To address the challenges of CIL or CIR, \ie, catastrophic forgetting~\cite{mccloskey1989catastrophic}, rehearsal-based methods have been explored. 
Rehearsal-based methods~\cite{rebuffi2017icarl, yoon2021online, buzzega2020dark, seo2024learning} are among the most widely studied approaches in CIL and CIR due to their effectiveness in mitigating forgetting of previous tasks, \ie, catastrophic forgetting~\cite{mccloskey1989catastrophic}.
These methods typically store a small subset~\cite{rebuffi2017icarl} or even the full dataset~\cite{prabhu2023online} of previous task exemplars in a memory buffer and replay them during training with new data. 
However, practical constraints such as data privacy concerns~\cite{zhang2023tag} and storage limitations can make rehearsal-based methods infeasible in real-world scenarios.
This leads to the development of rehearsal-free continual learning approaches~\cite{gao2022r, smith2021always, liu2022few} that aim to maintain model performance without relying on stored exemplars.

A promising variant within the rehearsal-free category involves leveraging open-source external data that can be easily accessed online. 
These data can be downloaded, utilized during training, and discarded afterwards, effectively circumventing the challenges related to storing sensitive or large amounts of data. Such an approach offers greater flexibility in model training and presents a valuable opportunity to enhance both the stability and plasticity of models in CIR. 
Therefore, developing strategies that maximize the benefits of open-source data becomes crucial in advancing the effectiveness of continual learning systems.

In this paper, we propose two components that utilize the external unlabeled data to effectively promote the stability and plasticity of the model. 
To encourage stability, we introduce \textbf{multi-level knowledge distillation (MLKD)}, which extracts learned knowledge from multiple previous models from various perspectives (\eg, features, logits) to compensate for the data distribution shift from the training data to the external unlabeled data. To enhance plasticity and facilitate faster learning, we use the unlabeled dataset to help the model extracting general features from the input, enabling it to identify relevant features from unseen classes and aiding in the classification of new classes. Thus, we apply \textbf{dynamic self-supervised loss (SSL)} in the continual training of the model.

Our multi-level knowledge distillation and dynamic self-supervised loss, built upon the baseline of local cross-entropy and feature replay, show significant performance improvements in the image classification task on the ImageNet-1K~\cite{deng2009imagenet} subset. 
Our method effectively utilizes unlabeled data to preserve knowledge, achieving 2nd place in the \textit{CVPR 2024 5th CLVISION Challenge}.

%% file: sec/2_method.tex
\section{Methodology}
\begin{figure*}
  \centering
    \includegraphics[width=\linewidth]{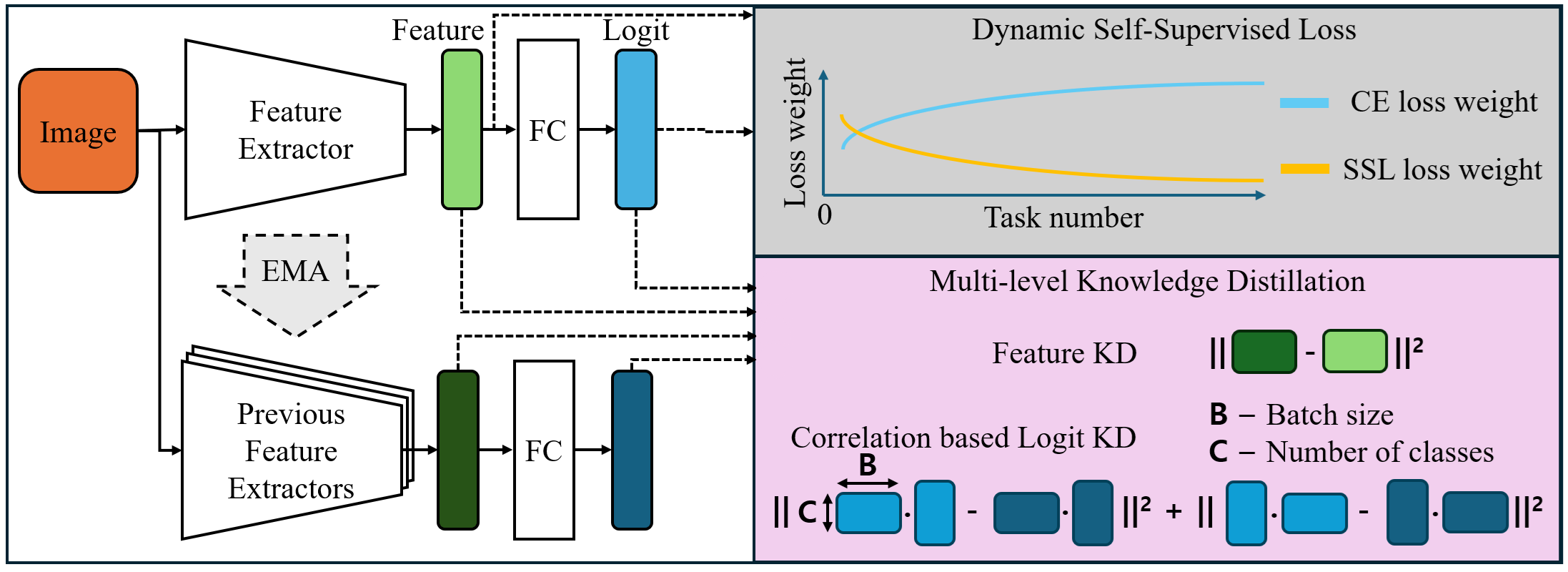}
  \caption{Overview of our method.}
  \label{fig:main}
\end{figure*}

We define the problem statement of class-incremental learning with repetition (CIR) utilizing an unlabeled data. 
We then explain how we address the plasticity and stability of the model through multi-level knowledge distillation (MLKD) and dynamic self-supervised learning (SSL). 
The overview of our method is illustrated in Figure~\ref{fig:main}.

\subsection{Problem Statement}
Let \( \mathcal{D}_l = \{ (x_i, y_i) \mid x_i \in X_l, y_i \in Y_l\}_{i=1}^N \) denote the labeled dataset,
which consists of \( N \) samples, each represented by an image-label pair \( (x_i, y_i) \) from an image set $X_l$ and a label space $Y_l = \{1,2,...,C\}$, and let \( \mathcal{D}_u = \{ x'_i \mid x'_i \in X_u\}_{i=1}^{N'} \) denote the unlabeled dataset with $N'$ samples $x'_i$. There exist a small memory buffer $M$ that can store relevant information about the previously learned tasks.

The objective of CIR is to maximize final accuracy in the image classification task on the test set with the data from the label space $Y_l$. 

In this paper, we denote the feature as the output of the feature extractor part of the model (\ie ResNet encoder), and the logit as the output of the prediction head of the model.

\subsection{Multi-Level Knowledge Distillation (MLKD)}

To alleviate catastrophic forgetting in CIR, it is crucial to distill knowledge from previous models. 
Extending from common knowledge distillation (KD) in continual learning~\cite{koh2022online, li2017learning}, our MLKD involves distilling knowledge from multiple perspectives across multiple previous models, utilizing both feature-level KD and logit-level KD.

For feature KD, we apply L2 loss to distill features from previous models. 
This helps ensure that the representations learned by the model remain consistent over time, preserving the knowledge embedded in these features. 
We linearly increase the weight of feature KD loss over tasks, as features from early stages may not contain much useful information and could hinder the learning of new knowledge.

We also apply KD using logits from previous models. 
However, naively distilling by instance-wise comparison is harmful, as the logits produced by the unlabeled data do not accurately reflect the knowledge of trained data due to data distribution discrepancies. 
To account for this distribution shift, we introduce a sophisticated KD mechanism, using Gram matrices to capture the correlations within batches and across classes~\cite{jin2023multi}, as follows:

\begin{dmath}
    L_{LogitKD}
    = \frac{1}{B} \sum_{k=1}^K ||G_{curr}^k - G_{prev}^k||^2 \\
    + \frac{1}{C} \sum_{k=1}^K ||M_{curr}^k - M_{prev}^k||^2.
\end{dmath}
The first term distills the correlation of instances within the batch, and the second term distills the correlation of classes.
$G$ and $M$ are Gram matrices, where for a logit batch $l \in \mathbb{R}^{B\text{ x }C}$ with batch size $B$ and the number of classes $C$, $G = l\cdot l^T$ and $M=l^T \cdot l$.

To extract a richer set of knowledge, we keep multiple previous models and mix their outputs for distillation. 
We maintain the trained model at the end of each task, and we select features and logits based on the confidence of the outputs. 
This multi-model approach enhances the robustness of knowledge transfer, compensating for the data distribution shift. 
Moreover, we gradually update all previous models through exponential moving average (EMA) with the current model, reflecting new knowledge in the distillation process and promoting the model's plasticity.

\subsection{Dynamic Self-Supervised Learning}

In addition to knowledge distillation, we incorporate self-supervised loss using the unlabeled dataset to foster the model's plasticity, enabling it to adapt to new concepts effectively. 
Self-supervised learning (SSL) encourages the model to extract general features that are useful across various tasks. 
This generalization capability is crucial for learning new concepts without significantly altering existing knowledge.

However, a naive application of SSL can degrade the performance of the primary task (e.g., image classification), as SSL does not directly align with the target task objectives.
To balance the influence of SSL on the model's learning process, we apply dynamic weighting to SSL, as follows:
\begin{dmath}
    L(X_l, X_u)
    = (1-0.1\cdot\alpha) L_{ACE}(X_l) + \alpha L_{SSL}(X_u),
\end{dmath}
where $\alpha = c\omega^t$, where $c$ and $\omega$ are hyperparameters, and $t$ is a task number. $\omega$ is set to less than 1, causing $\alpha$ to decreases over the tasks.
This ensures that the model remains focused on the primary task (\ie, image classification) while still benefiting from the generalization effects of SSL. 
 
\subsection{Final Loss}
We build our two components on a baseline commonly used in continual learning. Our final loss is as follows:
\begin{dmath}
L_{final}(X_l, X_u, M) =
\\ (1-0.1\cdot\alpha) L_{ACE}(X_l, M) + \alpha L_{SSL}(X_u) \\
+ \gamma L_{LC}(X_l) + \eta L_{der}(M) \\
+ \beta L_{featureKD}(X_u)  + \delta L_{logitKD}(X_u),
\end{dmath}
where $L_{ACE}$ is an asynchronous cross-entropy loss~\cite{caccia2021new} that compute loss only on classes in the batch, $L_{LC}$ is a logit constraint loss introduced in XDER~\cite{boschini2022class} to help discriminate the classes from different tasks, and we store features and logits of samples into a memory buffer $M$ for replay~\cite{buzzega2020dark}.

%% file: sec/3_experiment.tex
\section{Experiments}
\subsection{Challenge Settings}
The challenge assumes a continual learning setup with a stream of 50 experiences. 
The dataset is composed of a subset of ImageNet-1K~\cite{deng2009imagenet}, containing 130 classes related to birds, sea animals, and bugs. 
The labeled data stream contains data from 100 classes. For each experience introduced sequentially, 500 labeled data samples and 1000 unlabeled data samples are provided. 
The unlabeled data includes: 1) images of the same classes as the labeled data in the same experience, 2) images from within the 100 classes of the labeled data stream, and 3) random images from the whole dataset, depending on the challenge scenarios. 
The model architecture is ResNet-18~\cite{he2016deep}, with constraints of 8,000 MB GPU memory and a time limit of 600 minutes. 
The memory buffer is limited to storing only 200 exemplars, each of size less than 1,024 floating points.

\subsection{Implementation Details}
We implemented our method using PyTorch~\cite{paszke2017automatic} and Avalanche~\cite{lomonaco2021avalanche} libraries. 
We used Adam optimizer~\cite{kingma2014adam} with a learning rate of 4e-4. 
The batch size for labeled and unlabeled data was set to 64 and 100, respectively.
We chose rotation prediction~\cite{gidaris2018unsupervised} as the self-supervised loss.
The parameters were set as follows: $c = 0.5$, $\omega = 0.95$, $\gamma = 0.1$, 
$\eta = 0.4$, $\beta = 0.002t$, and $\delta = 0.1$.
For inference, we utilized model ensemble, averaging the predictions of the current model and one previous model.
All experiments were conducted on a single RTX 2080Ti GPU.
\subsection{Experiment Results}
\begin{table}
  \centering
  \begin{tabular}{@{}cc@{}}
    \toprule
    Method & Final Acc (\%) \\
    \midrule
    Fine-tuning & 5.82 \\
    Baseline & 19.54 \\
    Baseline + Dynamic SSL & 23.82 \\
    Baseline + MLKD & 39.82 \\
    \textbf{Baseline + Dynamic SSL + MLKD (Ours)} & \textbf{42.00} \\
    \bottomrule
  \end{tabular}
  \caption{Experiment results. }
  \label{tab:main_result}
\end{table}

The experiment results are presented in Table~\ref{tab:main_result}, showing the performance of different methods on the final accuracy. 
Our proposed method, combining multi-level knowledge distillation (MLKD) and dynamic self-supervised learning (SSL), achieves the highest accuracy of 42.00\%. 
The naive fine-tuning (FT) method results in the lowest accuracy at 5.82\%, indicating significant catastrophic forgetting. The baseline method improves the accuracy to 19.54\%, while adding dynamic SSL to the baseline further increases it to 23.82\%. 
Incorporating MLKD significantly boosts the accuracy to 39.82\%, demonstrating the effectiveness of our approach in preserving knowledge. 
Finally, our full method, which combines both dynamic SSL and MLKD, achieves the best performance.

\begin{table}
  \centering
  \begin{tabular}{@{}ccccc@{}}
    \toprule
    FKD & EMA & CLKD & MPM & Final Acc (\%) \\
    \midrule
    \checkmark & &&& 39.40 \\
    \checkmark & \checkmark &&& 40.84 \\
    \checkmark & \checkmark & \checkmark & & 41.08 \\
    \checkmark & \checkmark & \checkmark & \checkmark & 42.00 \\
    \bottomrule
  \end{tabular}
  \caption{Ablation study of MLKD. FKD, EMA, CLKD, and MPM refer to Feature KD, Exponential Moving Average, Correlation-based Logit KD, and Multiple Previous Models, respectively.}
  \label{tab:abl_multilevelkd}
\end{table}

The detailed ablation study of the components of MLKD is shown in Table~\ref{tab:abl_multilevelkd}.
Each component's contribution is analyzed by progressively adding them to the feature knowledge distillation (FKD) baseline. 
When only FKD is applied, the accuracy is 39.40\%. Adding exponential moving average (EMA) to FKD increases the accuracy to 40.84\%. 
Including correlation-based logit KD (CLKD) further improves the accuracy to 41.08\%. 
Finally, combining all components, including multiple previous models (MPM), achieves the highest accuracy of 42.00\%, confirming the synergistic effect of these components.

\begin{table}
  \centering
  \begin{tabular}{@{}cc@{}}
    \toprule
    Number of previous models & Final Acc (\%) \\
    \midrule
    1 previous model & 40.34 \\
    2 previous models & 41.56 \\
    3 previous models & 42.00 \\
    4 previous models & 41.84 \\
    \bottomrule
  \end{tabular}
  \caption{The number of previous models used in MLKD.}
  \label{tab:num_prevmodels}
\end{table}

Table~\ref{tab:num_prevmodels} explores the impact of the number of previous models used in MLKD. 
More previous models lead to better performance, but require more training time due to more number of model forwarding and EMA updating.
Using only one previous model results in an accuracy of 40.34\%. As we increase the number of previous models to two and three, the accuracy improves to 41.56\% and 42.00\%, respectively. However, adding a fourth previous model slightly decreases the accuracy to 41.84\%, suggesting that three previous models provide the optimal balance between retaining useful knowledge and avoiding redundancy with reasonable training time.

Table~\ref{tab:dynamic_ssl} compares the performance with and without dynamic weighting applied to the self-supervised loss. 
Without dynamic weighting, the model achieves an accuracy of 40.76\%. When dynamic weighting is applied, the accuracy improves to 42.00\%, demonstrating that dynamically adjusting the influence of SSL helps maintain the model's focus on the primary task while still benefiting from the generalization effects of SSL.

We also experimented with various self-supervised losses other than rotation prediction, as presented in Table~\ref{tab:ssl_choice}. Other recent state-of-the-art SSL methods perform worse than simple rotation prediction. We assume this is because these methods require a large batch size and a long training time to be effective. Thus, complying with the challenge`s restriction leads to sub-optimal performance when using these contrastive-based SSL methods.

\begin{table}
  \centering
  \begin{tabular}{@{}cc@{}}
    \toprule
    Method & Final Acc (\%) \\
    \midrule
    Without dynamic weighting & 40.76 \\
    \textbf{With dynamic weighting} & \textbf{42.00} \\
    \bottomrule
  \end{tabular}
  \caption{The performance difference with and without dynamic weighting to self-supervised loss.}
  \label{tab:dynamic_ssl}
\end{table}

\begin{table}
  \centering
  \begin{tabular}{@{}cc@{}}
    \toprule
    SSL Method & Final Acc (\%) \\
    \midrule
    \textbf{Rotation Prediction}~\cite{gidaris2018unsupervised} & 42.00 \\
    SimCLR~\cite{chen2020simple} & 39.10 \\
    SimSiam~\cite{chen2021exploring} & 39.46 \\
    VICReg~\cite{bardes2021vicreg} & 36.18 \\
    VICReg-ctr~\cite{garrido2022duality} & 37.26 \\
    \bottomrule
  \end{tabular}
  \caption{The performance difference with the choice of self-supervised loss.}
  \label{tab:ssl_choice}
\end{table}

%% file: sec/4_concolusion.tex
\section{Conclusion}
We propose multi-level knowledge distillation and adaptive self-supervised loss to promote stability and plasticity to utilize external unlabeled data instead of storing labeled data that the risk of privacy issues exist. 
These two approaches that are for maintaining the previous knowledge in the previously trained models and for accelerating the learning of new classes improve the existing continual learning baseline.
Various ablation studies are conducted to demonstrate the effectiveness of our method in class incremental with repetition, and we rank 2nd place on CVPR 5th CLVISION Challenge.

\section{Acknowledgment}
We acknowledge the EuroHPC Joint Undertaking for awarding this project access to the EuroHPC supercomputers MareNostrum5 at BSC, Spain; LEONARDO at CINECA, Italy; VEGA at IZUM, Slovenia; Karolina at IT4Innovations, Czech Republic; MeluXina at LuxProvide, Luxembourg; Discoverer at Sofia Tech Park, Bulgaria; and Deucalion at Minho Advanced Computing Centre, Portugal, under project IDs EHPC-DEV-2025D08-065 and EHPC-DEV-2025D08-088, through EuroHPC Development Access calls.

%% file: main.bbl
\begin{thebibliography}{27}
\providecommand{\natexlab}[1]{#1}
\providecommand{\url}[1]{\texttt{#1}}
\expandafter\ifx\csname urlstyle\endcsname\relax
  \providecommand{\doi}[1]{doi: #1}\else
  \providecommand{\doi}{doi: \begingroup \urlstyle{rm}\Url}\fi

\bibitem[Bardes et~al.(2021)Bardes, Ponce, and LeCun]{bardes2021vicreg}
Adrien Bardes, Jean Ponce, and Yann LeCun.
\newblock Vicreg: Variance-invariance-covariance regularization for self-supervised learning.
\newblock \emph{arXiv preprint arXiv:2105.04906}, 2021.

\bibitem[Boschini et~al.(2022)Boschini, Bonicelli, Buzzega, Porrello, and Calderara]{boschini2022class}
Matteo Boschini, Lorenzo Bonicelli, Pietro Buzzega, Angelo Porrello, and Simone Calderara.
\newblock Class-incremental continual learning into the extended der-verse.
\newblock \emph{IEEE transactions on pattern analysis and machine intelligence}, 45\penalty0 (5):\penalty0 5497--5512, 2022.

\bibitem[Buzzega et~al.(2020)Buzzega, Boschini, Porrello, Abati, and Calderara]{buzzega2020dark}
Pietro Buzzega, Matteo Boschini, Angelo Porrello, Davide Abati, and Simone Calderara.
\newblock Dark experience for general continual learning: a strong, simple baseline.
\newblock \emph{Advances in neural information processing systems}, 33:\penalty0 15920--15930, 2020.

\bibitem[Caccia et~al.(2021)Caccia, Aljundi, Asadi, Tuytelaars, Pineau, and Belilovsky]{caccia2021new}
Lucas Caccia, Rahaf Aljundi, Nader Asadi, Tinne Tuytelaars, Joelle Pineau, and Eugene Belilovsky.
\newblock New insights on reducing abrupt representation change in online continual learning.
\newblock \emph{arXiv preprint arXiv:2104.05025}, 2021.

\bibitem[Chen et~al.(2020)Chen, Kornblith, Norouzi, and Hinton]{chen2020simple}
Ting Chen, Simon Kornblith, Mohammad Norouzi, and Geoffrey Hinton.
\newblock A simple framework for contrastive learning of visual representations.
\newblock In \emph{International conference on machine learning}, pages 1597--1607. PMLR, 2020.

\bibitem[Chen and He(2021)]{chen2021exploring}
Xinlei Chen and Kaiming He.
\newblock Exploring simple siamese representation learning.
\newblock In \emph{Proceedings of the IEEE/CVF conference on computer vision and pattern recognition}, pages 15750--15758, 2021.

\bibitem[Deng et~al.(2009)Deng, Dong, Socher, Li, Li, and Fei-Fei]{deng2009imagenet}
Jia Deng, Wei Dong, Richard Socher, Li-Jia Li, Kai Li, and Li Fei-Fei.
\newblock Imagenet: A large-scale hierarchical image database.
\newblock In \emph{2009 IEEE conference on computer vision and pattern recognition}, pages 248--255. Ieee, 2009.

\bibitem[Gao et~al.(2022)Gao, Zhao, Ghanem, and Zhang]{gao2022r}
Qiankun Gao, Chen Zhao, Bernard Ghanem, and Jian Zhang.
\newblock R-dfcil: Relation-guided representation learning for data-free class incremental learning.
\newblock In \emph{European Conference on Computer Vision}, pages 423--439. Springer, 2022.

\bibitem[Garrido et~al.(2022)Garrido, Chen, Bardes, Najman, and Lecun]{garrido2022duality}
Quentin Garrido, Yubei Chen, Adrien Bardes, Laurent Najman, and Yann Lecun.
\newblock On the duality between contrastive and non-contrastive self-supervised learning.
\newblock \emph{arXiv preprint arXiv:2206.02574}, 2022.

\bibitem[Gidaris et~al.(2018)Gidaris, Singh, and Komodakis]{gidaris2018unsupervised}
Spyros Gidaris, Praveer Singh, and Nikos Komodakis.
\newblock Unsupervised representation learning by predicting image rotations.
\newblock \emph{arXiv preprint arXiv:1803.07728}, 2018.

\bibitem[He et~al.(2016)He, Zhang, Ren, and Sun]{he2016deep}
Kaiming He, Xiangyu Zhang, Shaoqing Ren, and Jian Sun.
\newblock Deep residual learning for image recognition.
\newblock In \emph{Proceedings of the IEEE conference on computer vision and pattern recognition}, pages 770--778, 2016.

\bibitem[Hemati et~al.(2023)Hemati, Cossu, Carta, Hurtado, Pellegrini, Bacciu, Lomonaco, and Borth]{hemati2023class}
Hamed Hemati, Andrea Cossu, Antonio Carta, Julio Hurtado, Lorenzo Pellegrini, Davide Bacciu, Vincenzo Lomonaco, and Damian Borth.
\newblock Class-incremental learning with repetition.
\newblock In \emph{Conference on Lifelong Learning Agents}, pages 437--455. PMLR, 2023.

\bibitem[Jin et~al.(2023)Jin, Wang, and Lin]{jin2023multi}
Ying Jin, Jiaqi Wang, and Dahua Lin.
\newblock Multi-level logit distillation.
\newblock In \emph{Proceedings of the IEEE/CVF Conference on Computer Vision and Pattern Recognition}, pages 24276--24285, 2023.

\bibitem[Karimireddy et~al.(2020)Karimireddy, Kale, Mohri, Reddi, Stich, and Suresh]{karimireddy2020scaffold}
Sai~Praneeth Karimireddy, Satyen Kale, Mehryar Mohri, Sashank Reddi, Sebastian Stich, and Ananda~Theertha Suresh.
\newblock Scaffold: Stochastic controlled averaging for federated learning.
\newblock In \emph{International conference on machine learning}, pages 5132--5143. PMLR, 2020.

\bibitem[Kingma and Ba(2014)]{kingma2014adam}
Diederik~P Kingma and Jimmy Ba.
\newblock Adam: A method for stochastic optimization.
\newblock \emph{arXiv preprint arXiv:1412.6980}, 2014.

\bibitem[Koh et~al.(2022)Koh, Seo, Bang, Song, Hong, Park, Ha, and Choi]{koh2022online}
Hyunseo Koh, Minhyuk Seo, Jihwan Bang, Hwanjun Song, Deokki Hong, Seulki Park, Jung-Woo Ha, and Jonghyun Choi.
\newblock Online boundary-free continual learning by scheduled data prior.
\newblock In \emph{The Eleventh International Conference on Learning Representations}, 2022.

\bibitem[Li and Hoiem(2017)]{li2017learning}
Zhizhong Li and Derek Hoiem.
\newblock Learning without forgetting.
\newblock \emph{IEEE transactions on pattern analysis and machine intelligence}, 40\penalty0 (12):\penalty0 2935--2947, 2017.

\bibitem[Liu et~al.(2022)Liu, Gu, Chi, Wang, Yu, Chen, and Tang]{liu2022few}
Huan Liu, Li Gu, Zhixiang Chi, Yang Wang, Yuanhao Yu, Jun Chen, and Jin Tang.
\newblock Few-shot class-incremental learning via entropy-regularized data-free replay.
\newblock In \emph{European Conference on Computer Vision}, pages 146--162. Springer, 2022.

\bibitem[Lomonaco et~al.(2021)Lomonaco, Pellegrini, Cossu, Carta, Graffieti, Hayes, Lange, Masana, Pomponi, van~de Ven, Mundt, She, Cooper, Forest, Belouadah, Calderara, Parisi, Cuzzolin, Tolias, Scardapane, Antiga, Amhad, Popescu, Kanan, van~de Weijer, Tuytelaars, Bacciu, and Maltoni]{lomonaco2021avalanche}
Vincenzo Lomonaco, Lorenzo Pellegrini, Andrea Cossu, Antonio Carta, Gabriele Graffieti, Tyler~L. Hayes, Matthias~De Lange, Marc Masana, Jary Pomponi, Gido van~de Ven, Martin Mundt, Qi She, Keiland Cooper, Jeremy Forest, Eden Belouadah, Simone Calderara, German~I. Parisi, Fabio Cuzzolin, Andreas Tolias, Simone Scardapane, Luca Antiga, Subutai Amhad, Adrian Popescu, Christopher Kanan, Joost van~de Weijer, Tinne Tuytelaars, Davide Bacciu, and Davide Maltoni.
\newblock Avalanche: an end-to-end library for continual learning.
\newblock In \emph{Proceedings of IEEE Conference on Computer Vision and Pattern Recognition}, 2021.

\bibitem[McCloskey and Cohen(1989)]{mccloskey1989catastrophic}
Michael McCloskey and Neal~J Cohen.
\newblock Catastrophic interference in connectionist networks: The sequential learning problem.
\newblock In \emph{Psychology of learning and motivation}, pages 109--165. Elsevier, 1989.

\bibitem[Paszke et~al.(2017)Paszke, Gross, Chintala, Chanan, Yang, DeVito, Lin, Desmaison, Antiga, and Lerer]{paszke2017automatic}
Adam Paszke, Sam Gross, Soumith Chintala, Gregory Chanan, Edward Yang, Zachary DeVito, Zeming Lin, Alban Desmaison, Luca Antiga, and Adam Lerer.
\newblock Automatic differentiation in pytorch.
\newblock 2017.

\bibitem[Prabhu et~al.(2023)Prabhu, Cai, Dokania, Torr, Koltun, and Sener]{prabhu2023online}
Ameya Prabhu, Zhipeng Cai, Puneet Dokania, Philip Torr, Vladlen Koltun, and Ozan Sener.
\newblock Online continual learning without the storage constraint.
\newblock \emph{arXiv preprint arXiv:2305.09253}, 2023.

\bibitem[Rebuffi et~al.(2017)Rebuffi, Kolesnikov, Sperl, and Lampert]{rebuffi2017icarl}
Sylvestre-Alvise Rebuffi, Alexander Kolesnikov, Georg Sperl, and Christoph~H Lampert.
\newblock icarl: Incremental classifier and representation learning.
\newblock In \emph{Proceedings of the IEEE conference on Computer Vision and Pattern Recognition}, pages 2001--2010, 2017.

\bibitem[Seo et~al.(2024)Seo, Koh, Jeung, Lee, Kim, Lee, Cho, Choi, Kim, and Choi]{seo2024learning}
Minhyuk Seo, Hyunseo Koh, Wonje Jeung, Minjae Lee, San Kim, Hankook Lee, Sungjun Cho, Sungik Choi, Hyunwoo Kim, and Jonghyun Choi.
\newblock Learning equi-angular representations for online continual learning.
\newblock In \emph{Proceedings of the IEEE/CVF Conference on Computer Vision and Pattern Recognition}, pages 23933--23942, 2024.

\bibitem[Smith et~al.(2021)Smith, Hsu, Balloch, Shen, Jin, and Kira]{smith2021always}
James Smith, Yen-Chang Hsu, Jonathan Balloch, Yilin Shen, Hongxia Jin, and Zsolt Kira.
\newblock Always be dreaming: A new approach for data-free class-incremental learning.
\newblock In \emph{Proceedings of the IEEE/CVF international conference on computer vision}, pages 9374--9384, 2021.

\bibitem[Yoon et~al.(2021)Yoon, Madaan, Yang, and Hwang]{yoon2021online}
Jaehong Yoon, Divyam Madaan, Eunho Yang, and Sung~Ju Hwang.
\newblock Online coreset selection for rehearsal-based continual learning.
\newblock \emph{arXiv preprint arXiv:2106.01085}, 2021.

\bibitem[Zhang et~al.(2023)Zhang, Xia, Liu, Xu, Hoang, Xing, Staples, Lu, and Zhu]{zhang2023tag}
Dawen Zhang, Boming Xia, Yue Liu, Xiwei Xu, Thong Hoang, Zhenchang Xing, Mark Staples, Qinghua Lu, and Liming Zhu.
\newblock Tag your fish in the broken net: A responsible web framework for protecting online privacy and copyright.
\newblock \emph{arXiv preprint arXiv:2310.07915}, 2023.

\end{thebibliography}
